\documentclass[11pt]{article}

\usepackage{acl}
\usepackage{times}
\usepackage{latexsym}
\usepackage{array}   
\usepackage{makecell} 
\usepackage{amsmath}   
\usepackage{multirow}
\usepackage{booktabs}
\usepackage{enumitem}
\usepackage{amsmath}
\usepackage{amssymb} 
\usepackage{xcolor}
\usepackage[T1]{fontenc}

\usepackage[utf8]{inputenc}

\usepackage{microtype}

\usepackage{inconsolata}
\usepackage{dblfloatfix}

\usepackage{graphicx}
\usepackage{subcaption}  
\usepackage{float} 
\usepackage{xcolor}
\usepackage[most]{tcolorbox}
\tcbuselibrary{listings,breakable}

\lstdefinestyle{promptstyle}{
  basicstyle=\ttfamily\small,
  breaklines=true,
  columns=fullflexible,
  keepspaces=true
}

\newtcolorbox{promptbox}[1]{
  enhanced,
  breakable,
  colback=gray!3,        
  colframe=gray!55,      
  boxrule=0.6pt,
  arc=3pt,
  left=6pt,right=6pt,top=6pt,bottom=6pt,
  title={#1},
  fonttitle=\bfseries,
  coltitle=black
}
\title{Patch the Distribution Mismatch: RL-Based Data Rewriting for Stable Downstream SFT}

\author{
  \textbf{Jiacheng Wang}\textsuperscript{1,*}, \textbf{Zhijie Liu}\textsuperscript{1,*},
  \textbf{Ping Jian}\textsuperscript{1,\textdagger}, \textbf{Zirong Chen}\textsuperscript{2}, \\
  \textbf{Ke Ren Liao}\textsuperscript{1}, \textbf{Zhen Yang}\textsuperscript{1}, \textbf{Zhongbin Guo}\textsuperscript{1} \\
  \textsuperscript{1}Beijing Institute of Technology, Beijing, China \quad
  \textsuperscript{2}Tsinghua University, Beijing, China \\
  \texttt{\{wangjc,pjian\}@bit.edu.cn} \\
  \textsuperscript{*}Equal contribution.\quad
  \textsuperscript{\textdagger}Corresponding author.
}

\begin{document}
\maketitle
\begin{abstract}
Large language models are commonly adapted to downstream tasks through supervised fine-tuning (SFT), but substantial distribution mismatch between downstream supervision and a model's generation distribution can intensify catastrophic forgetting. Data rewriting offers a data-centric way to narrow this mismatch before SFT. Existing methods, however, typically sample rewrites from a prompt-induced conditional distribution, which need not align with the backbone's natural question-answering generation distribution, and fixed templates can reduce output diversity. We formulate data rewriting as a policy-learning problem and train a lightweight LoRA rewriting policy with reinforcement learning. The policy optimizes question-answering-style distributional alignment and semantic diversity under a hard task-consistency gate, producing verified supervision for downstream SFT. Across three instruction-tuned backbones, the resulting models attain downstream gains broadly comparable to standard SFT while reducing degradation on non-downstream benchmarks in every evaluated setting. Additional experiments on logical reasoning and medical question answering provide preliminary evidence that a rewriting policy can be reused across domains for the same backbone.
\end{abstract}

\section{Introduction}

\label{introduction}
In recent years, Large Language Models (LLMs) have advanced rapidly~\citep{openai2024gpt4technicalreport} and demonstrated strong capabilities in tasks such as general dialogue~\citep{touvron2023llama2openfoundation}, information retrieval~\citep{sun2024chatgptgoodsearchinvestigating}, and reasoning~\citep{wei2023chainofthoughtpromptingelicitsreasoning}. 
However, when deployed in domain-specific or downstream scenarios, Supervised Fine-Tuning (SFT) on downstream data is still commonly required to improve performance on target tasks~\citep{ouyang2022traininglanguagemodelsfollow,dong2024abilitieslargelanguagemodels}. 
Despite its effectiveness, downstream SFT can inadvertently erode previously acquired general capabilities~\citep{luo2025empiricalstudycatastrophicforgetting,huang-etal-2024-mitigating} ($i.e.$ catastrophic forgetting)~\citep{li-etal-2024-revisiting}, especially under a substantial distribution shift between the downstream data and the model’s prior training data distribution~\citep{huang-etal-2025-selfaug}. 
In such cases, fine-tuning may bias the model toward the downstream distribution at the expense of other capabilities~\citep{kotha2024understandingcatastrophicforgettinglanguage}.
Recent work further observes that expert demonstrations can be low-likelihood under a model's generation distribution, making optimization more sensitive to distribution shift and potentially intensifying forgetting~\citep{zhang2025onpolicyrlmeetsoffpolicy,chen2025retainingdoingroleonpolicy}.

To reduce this distribution shift, a data-centric approach intervenes at the data source by rewriting the downstream training data before SFT~\citep{singh2024humandatascalingselftraining,zhang2025bestinstructiontuningdatafit}. 
The typical rewriting framework prompts the instruction-tuned base model $\pi_0$ with an input $x$, a reference solution $y^\star$, and a rewriting prompt $x_{\text{prompt}}$ to sample a rewrite $\tilde y$, where $\tilde y$ is task-consistent with $y^\star$ yet more in-distribution under $\pi_0$~\citep{yang2024selfdistillationbridgesdistributiongap,zhao2025mindgapdatarewriting}.
For example, SDFT~\citep{yang2024selfdistillationbridgesdistributiongap} follows a standard data-rewriting pipeline and trains on the rewritten data, whereas Mind the Gap~\citep{zhao2025mindgapdatarewriting} first lets the model attempt to solve each instance and rewrites expert demonstrations only for those it fails to solve.

Although data rewriting has shown promise in mitigating catastrophic forgetting, it still suffers from a key deficiency in distributional alignment.
Standard rewrites are typically sampled from a constrained, prompt-induced conditional distribution, $i.e.$ $\tilde y \sim \pi_0(\cdot \mid x, y^\star, x_{\text{prompt}})$, whereas downstream SFT more closely resembles QA-style completion conditioned only on $x$, corresponding to $\pi_0(\cdot \mid x)$.
Therefore, the rewrites may be more in-distribution only under the rewriting-prompt-induced conditional distribution, rather than genuinely closer to the QA-style generation distribution, and thus can only partially narrow the effective distribution gap.
This leads to two limitations: first, non-QA-style rewrites may introduce templated or unnatural phrasing, weakening the match between supervision and the model’s natural completion mode and degrading the efficiency and stability of downstream learning; second, sampling $\tilde y \sim \pi_0(\cdot \mid x, y^\star, x_{\text{prompt}})$ does not guarantee reducing the key gap relevant to $\pi_0(\cdot \mid x)$ (or $\pi_\theta(\cdot \mid x)$), and therefore its ability to mitigate catastrophic forgetting is likewise constrained.
Moreover, existing rewriting methods often rely on a small set of fixed prompt templates, which can further bias sampled rewrites toward templated and format-homogeneous patterns, inducing diversity collapse and under-representing many equally valid realizations~\citep{yun-etal-2025-price}.

Based on the above analysis, we model data rewriting as \emph{policy learning}, aiming to learn a rewriting policy that is consistent with the QA-style generation distribution and yields higher-quality rewrites.
In rewriting policy learning, task consistency, QA-style distributional alignment, and output diversity can all be formulated as automatically computable scalar feedback on sampled rewrites, making them naturally suitable as reward signals for policy optimization.
Compared to maximum-likelihood fitting under a constrained rewriting prompt, reinforcement learning can directly maximize the expected quality of sampled rewrites under these criteria, while incorporating non-differentiable verification procedures via hard gating or auxiliary shaping rewards; therefore, training the rewriting policy with RL is a natural choice in this setting.

Motivated by this perspective, we propose an RL-based data-rewriting agent $R_\phi$, which optimizes QA-style distributional alignment and diversity at the data source via automatically evaluable reward signals.
To reduce reward noise and stabilize policy learning, we treat task consistency as a hard constraint: we compute and optimize alignment/diversity rewards only for rewrites that pass a task-consistency verification gate.
Meanwhile, we view prompt-induced rewriting as a local deviation from the base model’s QA-style generation distribution and learn a lightweight low-rank residual “patch” via LoRA on top of a frozen base model $\pi_0$, enabling controllable local correction that suppresses policy drift and avoids over-correction~\citep{hu2022lora}.

\noindent Our contributions are summarized as follows:
\begin{itemize}[leftmargin=0pt,itemindent=\dimexpr\labelwidth+\labelsep\relax]
\item \textbf{Distribution Mismatch Insight:} We highlight a distribution mismatch in existing data-rewriting pipelines and propose an RL-based rewriting framework that learns to generate rewrites closer to the backbone’s QA-style generation distribution.

\item \textbf{Unified Objective:} We introduce a unified objective that enforces task consistency while jointly optimizing distributional alignment and diversity, thereby constructing a higher-quality rewritten dataset with stronger distributional consistency.

\item \textbf{Extensive Experiments:} Extensive experiments demonstrate that our method achieves downstream performance comparable to standard SFT, while substantially mitigating catastrophic forgetting on general-domain benchmarks.
\end{itemize}

\noindent Our code is publicly available at
\url{https://github.com/Jiacheng-31/Patch-the-Prompt-Gap}.

\section{Related Work}

\begin{figure*}[t]
  \centering
  \includegraphics[width=1\linewidth]{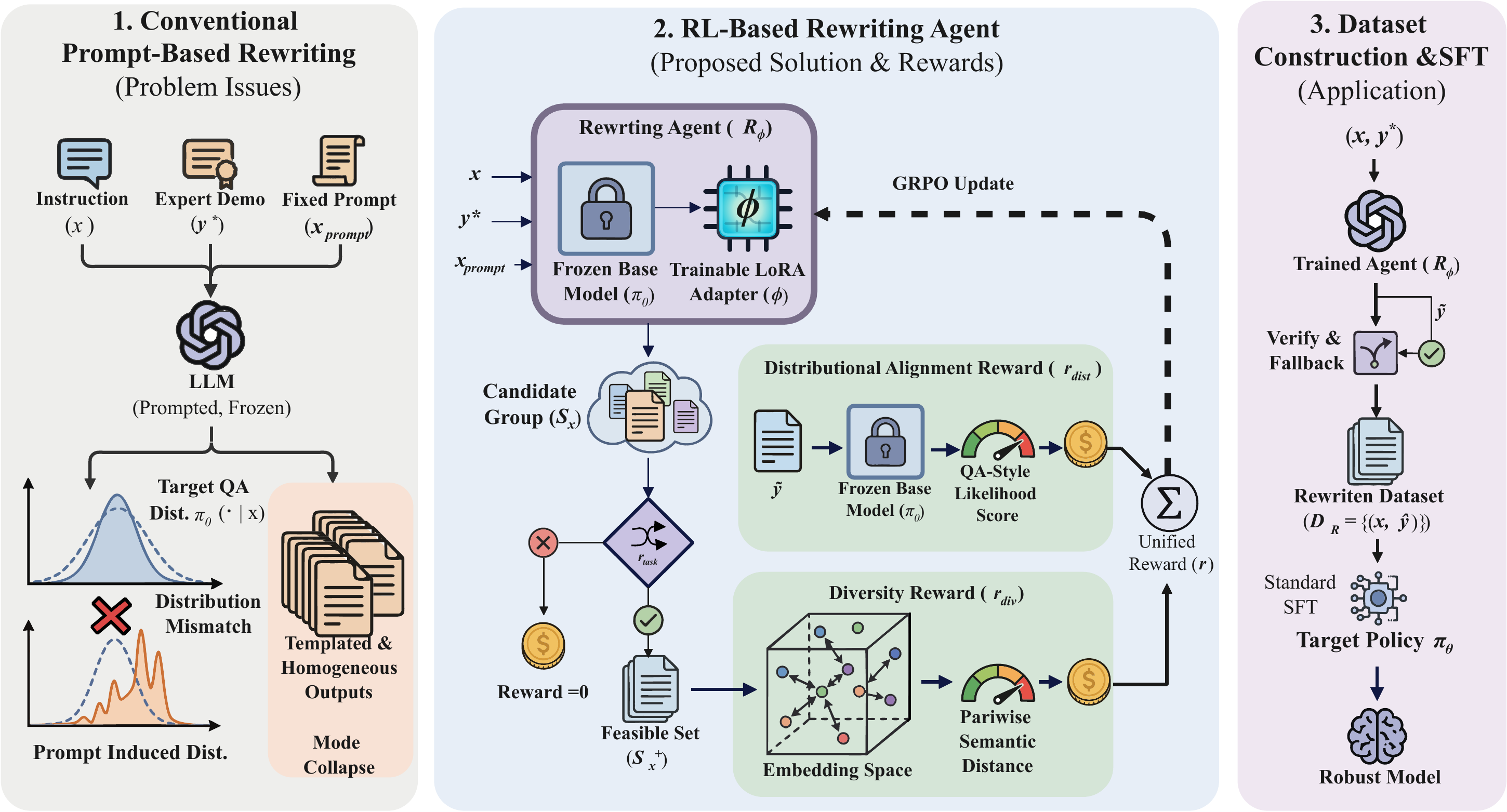}
  \caption{The framework of rewriting agent.}
  \label{fig:fig1}
\end{figure*}

\paragraph{Catastrophic Forgetting and Optimization.}
Catastrophic forgetting is widely observed when adapting large language models (LLMs) to domain-specific or downstream data, where downstream gains often come at the expense of general capabilities~\citep{luo2025empiricalstudycatastrophicforgetting,kang2025selfevolvingllmscontinualinstruction}.
A traditional line of work mitigates forgetting via rehearsal-based continual learning and pseudo-rehearsal, replaying data from earlier stages to preserve prior abilities~\citep{lopezpaz2022gradientepisodicmemorycontinual,wang-etal-2024-inscl,huang-etal-2024-mitigating,wang2025sers_posterpage}.

More recently, a growing line of work stabilizes post-training by introducing conservative-update mechanisms, including DFT-style anchoring~\citep{wu2025generalizationsftreinforcementlearning} and proximal or trust-region constraints~\citep{zhu2025anchoredsupervisedfinetuning,zhu2025proximalsupervisedfinetuning}, to bound per-step updates and suppress policy drift.
However, while these methods improve stability, they can also restrict adaptation under distribution shift, reflecting a practical stability--plasticity tension~\citep{dohare2024lossplasticity}.
In contrast, data rewriting offers a data-centric alternative: it proactively narrows the distribution gap between expert demonstrations and the current policy by rewriting supervision targets, preserving expert knowledge while making downstream supervision more in-distribution, thereby stabilizing training and reducing forgetting~\citep{yang2024selfdistillationbridgesdistributiongap,zhao2025mindgapdatarewriting}.

\paragraph{Diversity Collapse Induced by Formatting and Alignment.}
Recent studies show that structured templates and rigid formatting constraints can substantially suppress output diversity in instruction-tuned LLMs, leading to diversity/mode collapse~\citep{yun-etal-2025-price,2025_Xiao}.
Related analyses further attribute alignment-induced mode collapse to typicality bias in preference data, suggesting that alignment or template-like prompting can systematically contract the effective output space~\citep{zhang2025verbalizedsamplingmitigatemode,li2024preserving_diversity_sft}.

\section{Method}

\subsection{Problem Setup}
Consider a downstream supervised dataset $\mathcal D_E=\{(x_i,y_i^\star)\}_{i=1}^N$ and a target policy $\pi_\theta$ initialized from $\pi_0$.
Standard SFT on $\mathcal D_E$ can be optimization-unstable under distribution mismatch, exacerbating catastrophic forgetting; see Appendix~\ref{app:sft_instability}.
We assume access to the instruction-tuned backbone and downstream adaptation data, but not to the backbone's original training corpus or a representative general-domain replay corpus.
Under this setting, we construct an optimized training dataset that narrows the mismatch at the data level, with the aim of stabilizing SFT updates and mitigating forgetting.

\subsection{Framework}
To mitigate distribution-mismatch-induced instability in downstream SFT, we adopt a data-centric intervention that rewrites supervision before fine-tuning.
While prior data-rewriting pipelines~\citep{yang2024selfdistillationbridgesdistributiongap,zhao2025mindgapdatarewriting} largely rely on prompt-driven heuristics with a small set of fixed templates, we instead train a dedicated data-rewriting agent $R_{\phi}$ to perform rewriting.
The goal is to produce higher-quality supervision that preserves expert knowledge, better matches the base model’s natural QA-style distribution $\pi_0(\cdot\mid x)$, and explicitly avoids mode collapse by encouraging diverse yet acceptable realizations.

\paragraph{Stage I: RL training of the rewriting agent.}
We learn a parameter-efficient rewriting policy $R_\phi$ on top of a frozen $\pi_0$ via LoRA adapters~\citep{hu2022lora}.
Given an input $x$, its expert demonstration $y^\star$, and a rewriting prompt $x_{\text{prompt}}$, the agent generates a rewrite
\begin{equation}
\tilde y \sim R_\phi(\cdot \mid x, y^\star, x_{\text{prompt}})\, .
\end{equation}
We train $R_\phi$ with on-policy RL using automatically evaluable signals that jointly target (i) task consistency, (ii) distributional alignment to $\pi_0(\cdot\mid x)$ (conditioning on $x$ only), and (iii) diversity as an anti-collapse regularizer; the unified reward is detailed in Sec.~\ref{sec:reward}.

\paragraph{Stage II: dataset construction and downstream SFT.}
We apply the trained $R_\phi$ to construct a rewritten dataset $\mathcal D_R$ (with verification and fallback), and then perform standard SFT of $\pi_\theta$ (initialized from $\pi_0$) on $\mathcal D_R$ to obtain the final model. Details are given in Sec.~\ref{Downstream-SFT}.

\subsection{Parameter-Efficient Rewriting Policy and GRPO Optimization}
We employ the frozen instruction-tuned base model $\pi_0$ as the backbone and parameterize the rewriting policy $R_\phi$ with LoRA adapters, training only a small set of adapter parameters $\phi$~\citep{hu2022lora}. 
This parameter-efficient design is well-suited for rewriting, which mainly requires lightweight calibration of existing generation behaviors rather than learning a new distribution from scratch.
Moreover, freezing the backbone and restricting updates to low-rank adapters provides a capacity-limited ``patch'' over $\pi_0$, which helps bias learning toward targeted adjustments and empirically reduces unnecessary drift.

\noindent For optimization, we update $R_\phi$ using Group Relative Policy Optimization (GRPO)~\citep{shao2024deepseekmathpushinglimitsmathematical}, a PPO-style on-policy policy-gradient method that estimates the baseline from groups of sampled outputs, eliminating the need for a separate value network.
The overall objective is:
\begin{equation}
\max_{\phi}\ \mathbb{E}_{(x,y^\star)\sim \mathcal{D}_E,\ \tilde y \sim R_\phi(\cdot\mid x,y^\star,x_{\text{prompt}})}
\big[r(x,y^\star,\tilde y)\big] \, .
\end{equation}

\subsection{Unified Objective: Task Consistency, Distributional Alignment, and Diversity}
\label{sec:reward}
To jointly optimize task consistency, distributional alignment, and diversity, we define a unified objective.
For each expert sample $(x,y^\star)$, we sample a group of $K$ candidate rewrites
$\tilde y^{(k)} \sim R_\phi(\cdot \mid x,y^\star,x_{\text{prompt}})$
to form the indexed multiset $\mathcal S_x=\{\tilde y^{(k)}\}_{k=1}^{K}$; repeated texts remain distinct
candidates because their sample indices determine group-relative credit assignment.
For any candidate $\tilde y$, we define three reward components: $r_{\text{task}}(x,y^\star,\tilde y)$,
$r_{\text{dist}}(x,\tilde y)$, and $r_{\text{div}}(x,\tilde y;\mathcal S_x^+)$ to measure task consistency,
distributional alignment, and diversity, respectively, where
$\mathcal S_x^{+}=\{\tilde y^{(k)}: k\in I_x^+\}$ is the indexed feasible sub-multiset and
$I_x^+=\{k:r_{\text{task}}(x,y^\star,\tilde y^{(k)})=1\}$.
We treat $r_{\text{task}}\in\{0,1\}$ as a hard feasibility gate:
the distributional-alignment and diversity rewards are applied only if a rewrite passes the task-consistency check;
otherwise, these auxiliary metrics are masked out.
Accordingly, we use the following gated total reward:
\begin{equation}
\begin{gathered}
r(x,y^\star,\tilde y)=
r_{\text{task}}(x,y^\star,\tilde y)+\\
r_{\text{task}}(x,y^\star,\tilde y)\cdot\bigl(\lambda_{\text{dist}}\, r_{\text{dist}}(x,\tilde y)
+\lambda_{\text{div}}\, r_{\text{div}}(x,\tilde y;\mathcal S_x^+)\bigr)
\end{gathered}
\end{equation}
This gating ensures that invalid rewrites ($r_{\text{task}}=0$) receive zero reward and that auxiliary rewards are
skipped when infeasible, preventing misleading shaping signals while improving training stability and computational
efficiency.

\subsubsection{Task Consistency Reward}
The task consistency reward $r_{\text{task}}(x,y^\star,\tilde y)$ is a binary gate that checks whether a rewritten sample $\tilde y$ satisfies: (i) final-answer correctness and (ii) reasoning validity.
We adopt a coarse-to-fine verification scheme: we first use a low-cost, rule-based verifier to check the final answer; only when the answer is deemed correct do we invoke \textit{DeepSeek-V3.2} as an LLM judge to assess reasoning consistency and logical soundness. This avoids unnecessary evaluation on clearly incorrect samples and reduces cases where the answer is correct but the reasoning is unreliable~\citep{zheng2023judgingllmasajudgemtbenchchatbot,liu2023gevalnlgevaluationusing}.
Formally, we define
\begin{equation}
v_{\text{ans}}(x,y^\star,\tilde y)\in\{0,1\},
v_{\text{rea}}(x,y^\star,\tilde y)\in\{0,1\},
\end{equation}
where $v_{\text{ans}}$ denotes the rule-based judgment of final-answer correctness, and $v_{\text{rea}}$ denotes the LLM judge's assessment of reasoning validity (computed only when $v_{\text{ans}}=1$).
We define the task-consistency reward as
\begin{equation}
r_{\text{task}}(x,y^\star,\tilde y)
=
v_{\text{ans}}(x,y^\star,\tilde y)\cdot v_{\text{rea}}(x,y^\star,\tilde y)
\in\{0,1\}.
\end{equation}


\subsubsection{Distributional Alignment Reward}
%
To encourage rewrites that are in-distribution under standard QA-style generation, we score each rewrite using the frozen base model under the QA condition $\pi_0(\cdot \mid x)$ and reward higher generatability ($i.e.$ higher likelihood) under this distribution.
Specifically, we compute the length-normalized negative log-likelihood (NLL) of $\tilde y$:
\begin{equation}
\ell_{\text{dist}}(x,\tilde y)
=
-\frac{1}{|\tilde y|}
\sum_{t=1}^{|\tilde y|}
\log \pi_0(\tilde y_t \mid x, \tilde y_{<t}) \, .
\end{equation}
We then map it to a bounded, monotonic reward via group-wise normalization.
For a nonempty feasible candidate sub-multiset $\mathcal S_x^{+}$, we compute the mean and standard deviation of
$\ell_{\text{dist}}$ over its indexed members, denoted as $\mu_x$ and $\sigma_x$, and define the normalized score.
If $\mathcal S_x^{+}$ is empty, we set $r_{\text{dist}}=0$ for every candidate in the group.
\begin{equation}
\hat{\ell}_{\text{dist}}(x,\tilde y)
=\frac{\ell_{\text{dist}}(x,\tilde y)-\mu_x}{\sigma_x+\epsilon}\, ,
\end{equation}
where $\epsilon$ is a small constant for numerical stability.
We then apply a bounded, monotonic mapping:
\begin{equation}
r_{\text{dist}}(x,\tilde y)
=\frac{1}{1+\exp\!\left(\hat{\ell}_{\text{dist}}(x,\tilde y)\right)}.
\end{equation}
This group-wise normalization improves numerical resolution by removing prompt-dependent scale in NLL values, while preserving the within-group ordering.
It is also aligned with common stability practices in GRPO-style optimization, where reward whitening is used to stabilize learning dynamics.
Consequently, the policy is encouraged to generate rewrites that are more in-distribution under the base model's QA-style generation $\pi_0(\cdot\mid x)$.

\subsubsection{Diversity Reward}
To mitigate mode collapse and template-induced homogeneity, we encourage \emph{semantic diversity} among \emph{feasible} (task-consistent) rewrites for the same input $x$, and compute diversity only on $\mathcal S_x^{+}$ to avoid noise from invalid samples.
For brevity, when $(x,y^\star)$ is clear from context, we write $r_{\text{task}}(\tilde y)=r_{\text{task}}(x,y^\star,\tilde y)$, and let $m=|I_x^+|$.
When $m<2$, semantic diversity is ill-defined, so we set $r_{\text{div}}=0$; when $m\ge2$, we map each indexed candidate $\tilde y^{(k)}$, $k\in I_x^+$, into a semantic space with an embedding function $f(\cdot)$ and normalize it as
$e^{(k)}=\frac{f(\tilde y^{(k)})}{\|f(\tilde y^{(k)})\|}$.
We define $D(\mathcal S_x^+)=0$ when $m\le1$ and use the pairwise definition below when $m\ge2$.
We define the pairwise semantic distance as
\begin{equation}
d\!\left(e^{(i)},e^{(j)}\right)
=
\frac{1-\cos\!\left(e^{(i)},e^{(j)}\right)}{2}\in[0,1].
\end{equation}
The set-level semantic diversity is
\begin{equation}
D(\mathcal S_x^{+})
=
\frac{2}{m(m-1)}
\sum_{\substack{i,j\in I_x^+\\ i<j}}
d\!\left(e^{(i)},e^{(j)}\right).
\end{equation}

To provide fine-grained credit assignment, we define a \emph{marginal contribution} diversity reward for each feasible candidate.
For $k\in I_x^+$, let $\mathcal S_{x,-k}^{+}$ be the indexed feasible sub-multiset obtained by removing only candidate index $k$ from $\mathcal S_x^{+}$, and define its marginal contribution as
\begin{equation}
\Delta^{(k)}(\mathcal S_x^{+})
=
D(\mathcal S_x^{+}) - D(\mathcal S_{x,-k}^{+}).
\end{equation}
Accordingly, the diversity reward is
\begin{equation}
r_{\text{div}}^{(k)}
=
\mathbf{1}\!\left[r_{\text{task}}(\tilde y^{(k)})=1\right]\cdot
\big[\Delta^{(k)}(\mathcal S_x^{+})\big]_+ ,
\end{equation}
where $[z]_+=\max(0,z)$.
This marginal formulation encourages each feasible rewrite to contribute novel semantic content to the candidate set, directly discouraging template-induced homogeneity while avoiding spurious diversity signals from invalid rewrites.

\subsection{Rewriting Dataset Construction and Downstream SFT}
\label{Downstream-SFT}
After training the rewriting policy $R_\phi$, we construct the rewritten dataset via a \textsc{Generate--Verify--Fallback} pipeline.
For each expert sample $(x,y^\star)\in\mathcal D_E$, we first sample a rewrite
\begin{equation}
\tilde y \sim R_\phi(\cdot \mid x, y^\star, x_{\text{prompt}})\, .
\end{equation}
We then re-apply the same task-consistency check used during training, $r_{\text{task}}(x,y^\star,\tilde y)$.
If the check passes ($r_{\text{task}}=1$), we adopt $\tilde y$ as the supervision target; otherwise ($r_{\text{task}}=0$), we fall back to the original expert demonstration $y^\star$ to avoid losing supervision due to rewriting failures.
Formally, we set
\begin{equation}
\hat y=
\begin{cases}
\tilde y, & r_{\text{task}}(x,y^\star,\tilde y)=1,\\
y^\star, & \text{otherwise}.
\end{cases}
\end{equation}
Collecting $\{(x_i,\hat y_i)\}_{i=1}^{N}$ yields the rewritten training set $\mathcal D_R$.
Finally, we perform standard supervised fine-tuning of the target policy $\pi_\theta$ (initialized from $\pi_0$) on $\mathcal D_R$ using maximum likelihood training, obtaining the final model.

\section{Experiments}

\subsection{Experimental Setup}

\begin{table*}[t]
  \centering
  \setlength{\tabcolsep}{6pt}
  \renewcommand{\arraystretch}{1}
  \begin{tabular}{@{}l c c c c c@{}}
    \hline
    \textbf{Model}
      & \textbf{MathAvg}
      & \textbf{Math$\uparrow$ (\%)}
      & \textbf{GeneralAvg}
      & \textbf{Gen$\downarrow$ (\%)}
      & \textbf{OverallAvg} \\
    \hline
    \textbf{Llama-3.2-1B-Instruct} & 12.42 & -- & 27.34 & -- & 19.88 \\
    \quad + SFT & 15.40 & \textbf{$\uparrow$23.99} & 22.60 & $\downarrow$17.34 & 19.00 \\
    \quad + SDFT & 13.80 & $\uparrow$11.11 & 23.80 & $\downarrow$12.95 & 18.80 \\
    \quad + Mind the Gap & 12.92 & $\uparrow$4.03 & 24.41 & $\downarrow$10.72 & 18.67 \\
    \quad + GRPO & 12.88 & $\uparrow$3.70 & 25.85 & $\downarrow$5.45 & 19.37 \\
    \quad + Rewriting-agent & 15.10 & $\uparrow$21.58 & 25.92 & \textbf{$\downarrow$5.19} & \textbf{20.51} \\
    \hline

    \textbf{Llama-3.2-3B-Instruct} & 18.90 & -- & 45.21 & -- & 32.06 \\
    \quad + SFT & 21.81 & \textbf{$\uparrow$15.40} & 37.26 & $\downarrow$17.58 & 29.54 \\
    \quad + SDFT & 19.84 & $\uparrow$4.97 & 40.32 & $\downarrow$10.82 & 30.08 \\
    \quad + Mind the Gap & 20.30 & $\uparrow$7.41 & 40.44 & $\downarrow$10.55 & 30.37 \\
    \quad + GRPO & 19.83 & $\uparrow$4.92 & 43.18 & $\downarrow$4.49 & 31.51 \\
    \quad + Rewriting-agent & 21.12 & $\uparrow$11.75 & 43.17 & \textbf{$\downarrow$4.51} & \textbf{32.15} \\
    \hline

    \textbf{Mistral-7B-Instruct-v0.3} & 10.61 & -- & 40.53 & -- & 25.57 \\
    \quad + SFT & 15.37 & $\uparrow$44.88 & 32.76 & $\downarrow$19.17 & 24.06 \\
    \quad + SDFT & 13.77 & $\uparrow$29.78 & 35.36 & $\downarrow$12.76 & 24.57 \\
    \quad + Mind the Gap & 14.60 & $\uparrow$37.61 & 34.43 & $\downarrow$15.05 & 24.52 \\
    \quad + GRPO & 13.55 & $\uparrow$27.71 & 36.28 & $\downarrow$10.49 & 24.92 \\
    \quad + Rewriting-agent & 16.47 & \textbf{$\uparrow$55.23} & 37.55 & \textbf{$\downarrow$7.35} & \textbf{27.01} \\
    \hline
  \end{tabular}
  \caption{
    Math$\uparrow$ denotes the relative MathAvg improvement over the instruct-tuned base within the same block:
    $(M-M_{\text{base}})/M_{\text{base}}\times 100$.
    Gen$\downarrow$ denotes the relative GeneralAvg drop:
    $(G_{\text{base}}-G)/G_{\text{base}}\times 100$.
    OverallAvg $=(\mathrm{MathAvg}+\mathrm{GeneralAvg})/2$.
  }
  \label{tab:summary-results}
\end{table*}

Following the Mind the Gap~\citep{zhao2025mindgapdatarewriting} setting, we treat mathematical reasoning as the downstream capability to improve, and quantify catastrophic forgetting via performance changes on general-domain benchmarks.
Consistent with our problem setup, all methods use only the instruction-tuned backbone and downstream adaptation data; we do not assume access to the backbone's original training corpus or a representative general-domain replay corpus.
\paragraph{Datasets.}
\label{exp:datasets}
For training data, we use two widely adopted math corpora, \texttt{NuminaMath-CoT}~\citep{numina_math_datasets} and \texttt{OpenMathReasoning}~\citep{moshkov2025aimo2}.
We randomly sample from the merged pool, filter out overlong examples ($>8192$ tokens), and obtain 100K instances in total, split evenly into 50K for GRPO-based rewriting-agent training and 50K for downstream SFT.
For downstream math evaluation, we report results on \texttt{Math500}~\citep{lightman2023lets}, \texttt{MinervaMath}~\citep{2206.14858}, \texttt{AMC23}~\citep{yao2025fansformalanswer}, \texttt{AGIEval-Math}~\citep{zhong2023agieval}, and \texttt{IMO-Bench}~\citep{luong-etal-2025-towards}.
To assess out-of-domain generalization and quantify catastrophic forgetting, we further evaluate on \texttt{MMLU}~\citep{hendryckstest2021}, \texttt{MMLU-Pro}~\citep{wang2024mmlu}, and \texttt{AGIEval}~\citep{zhong2023agieval} with math-related subsets removed.
Detailed dataset descriptions are provided in Appendix~\ref{app:dataset}.

\paragraph{Backbone Models.}
We experiment with instruction-tuned backbones of different scales and training recipes:
\texttt{Llama-3.2-1B-Instruct}, \texttt{Llama-3.2-3B-Instruct}~\citep{grattafiori2024llama3herdmodels} and \texttt{Mistral-7B-Instruct-v0.3}~\citep{jiang2023mistral7b}.
This allows us to test whether the proposed rewriting-agent framework is robust across model sizes and families.


\paragraph{Baselines.}
Under the same backbone $\pi_0$, the same task-consistency verifier, and identical downstream SFT hyperparameters and data split, we compare:
(1)\texttt{Vanilla SFT} (SFT on the original corpus),
(2)\texttt{SDFT}~\citep{yang2024selfdistillationbridgesdistributiongap} (template-based rewrite + verify + fallback),
(3)\texttt{Mind the Gap}~\citep{zhao2025mindgapdatarewriting} (self-solve, then rewrite failed cases + verify + fallback),
(4)\texttt{GRPO} (on-policy RL fine-tuning of the target model with the same LoRA parameter budget as the rewriting agent),
and (5)\texttt{Rewriting-agent (Ours)} (learned policy rewriting + the same verify/fallback rule).
Detailed descriptions are provided in Appendix~\ref{app:baselines}.

\paragraph{Training Details.}
For downstream SFT, we use \texttt{LLaMA-Factory}~\citep{zheng2024llamafactory} with the AdamW optimizer, a global batch size of 128, and fine-tune for 2 epochs.
We use a learning rate of $5\times10^{-6}$ for \texttt{Mistral-7B-Instruct-v0.3}, and $7\times10^{-6}$ for both \texttt{Llama-3.2-1B-Instruct} and \texttt{Llama-3.2-3B-Instruct}.
For training the rewriting agent, we implement GRPO optimization with \texttt{verl}~\citep{sheng2024hybridflow}.
We freeze the backbone $\pi_0$ and update only the LoRA adapter parameters.
We use a global batch size of 512 prompts and sample $K{=}10$ candidate rewrites per prompt.
Further details are provided in Appendix~\ref{app:training-details}.


\begin{figure*}[t]
  \centering
  \includegraphics[width=\textwidth]{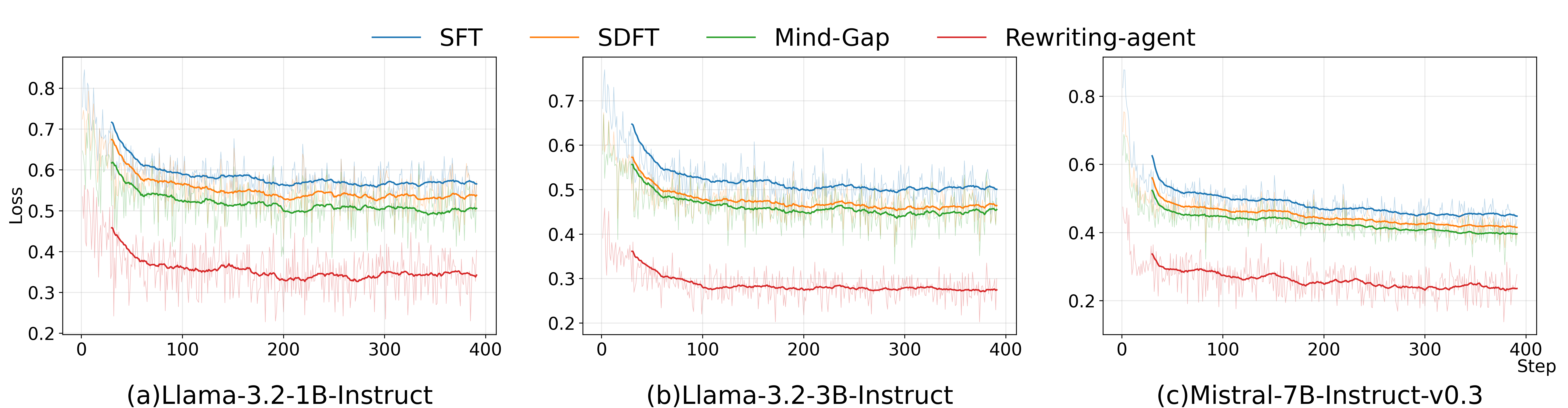}
  \caption{Downstream SFT training loss over training steps for three backbones and four methods.}
  \label{fig:loss}
\end{figure*}

\subsection{Main Results}

Table~\ref{tab:summary-results} summarizes the trade-off between downstream math gains and out-of-distribution generalization retention across three instruction-tuned backbones.
Overall, \texttt{Vanilla SFT} yields strong math gains, but it also incurs severe degradation on general-domain benchmarks, indicating pronounced catastrophic forgetting.
Fixed-prompt self-rewriting baselines (\texttt{SDFT} and \texttt{Mind the Gap}) partially mitigate this drop, but generally achieve smaller math improvements, suggesting that heuristic prompting alone is insufficient to jointly optimize task improvement and distributional alignment.
\texttt{GRPO} (direct on-policy RL fine-tuning) exhibits reduced forgetting comparable to our method, but yields limited downstream gains (e.g., only $+4.92\%$ on \texttt{Llama-3.2-3B}), indicating that pure on-policy exploration provides a weak learning signal on hard reasoning tasks.

In contrast, our \texttt{Rewriting-agent} achieves the best OverallAvg across all three backbones and consistently improves the gain--forgetting trade-off: it attains math gains broadly comparable to \texttt{Vanilla SFT} while exhibiting substantially smaller generalization drops.
For example, on \texttt{Mistral-7B-Instruct-v0.3}, \texttt{Rewriting-agent} exceeds SFT in math gains ($+55.23\%$ vs.\ $+44.88\%$) while reducing the generalization drop from $19.17\%$ to $7.35\%$.
Compared with \texttt{GRPO}, our method achieves much stronger downstream adaptation (e.g., $+11.75\%$ vs.\ $+4.92\%$ on \texttt{Llama-3.2-3B}) with a comparable level of generalization retention.
Overall, these results support the value of learning to rewrite supervision under explicit objectives of task consistency, QA-style generatability, and diversity for improving the gain--forgetting trade-off in the evaluated settings.
Detailed per-benchmark results are reported in Appendix~\ref{app:combined-results}.

\paragraph{Cost and reuse.}
On \texttt{Llama-3.2-3B-Instruct}, training the rewriting agent costs 38.226 A100-80G GPU-hours; constructing the rewritten dataset and downstream SFT require 1.472 and 5.811 GPU-hours, respectively (Appendix~\ref{app:cost}).
The agent-training cost is incurred once per backbone and can be amortized when the same agent is reused across downstream domains.
Our transfer experiments support reuse across math, logic, and medical QA for the same backbone; they do not establish cost-free transfer across different backbones.



\subsection{Analysis}


\paragraph{Why does the Rewriting-agent reduce forgetting?}
We attribute the improved retention to the stronger alignment between our rewritten data and the backbone’s QA-style generation distribution.
Specifically, Figure~\ref{fig:loss} shows that, across backbones, downstream SFT on \texttt{Rewriting-agent} targets starts from a noticeably lower training loss and converges to a lower and more stable plateau, outperforming \texttt{SFT} and fixed-prompt self-rewriting baselines (\texttt{SDFT} and \texttt{Mind the Gap}).
This suggests that the rewritten data are easier to fit under the standard QA-style maximum-likelihood objective, which is consistent with more stable optimization and reduced reliance on abrupt, large-magnitude updates; such updates can over-shift the policy toward the downstream distribution and exacerbate catastrophic forgetting.

\begin{table}[t]
  \centering
  \small
  \setlength{\tabcolsep}{4pt}
  \renewcommand{\arraystretch}{1.1}
  \begin{tabular}{lccc}
    \toprule
    \textbf{Method} & \textbf{Llama-1B} & \textbf{Llama-3B} & \textbf{Mistral-7B} \\
    \midrule
    \texttt{SDFT}          & 53.39 & 66.33 & 69.28 \\
    \texttt{Mind the Gap}  & 53.47 & 67.42 & 69.74 \\
    \texttt{Rewriting-agent} & 60.19 & 74.96 & 77.43 \\
    \bottomrule
  \end{tabular}
  \caption{Task-consistency yield (\%): fraction of rewrites passing $r_{\text{task}}$ for each method and backbone.}
  \label{tab:yield}
\end{table}

\paragraph{Task-consistency yield.}
We report the pass rate of the task-consistency gate $r_{\text{task}}$ during rewriting in Table~\ref{tab:yield}.
Compared to fixed-prompt rewriting, the learned rewriting policy substantially increases the fraction of feasible rewrites, thereby reducing reliance on fallback to the original expert demonstrations.
As a result, the constructed rewritten dataset $\mathcal D_R$ exhibits higher supervision quality and stronger distributional consistency, which in turn improves generalization retention and leads to less catastrophic forgetting.

\begin{figure}[t]
  \centering
  \includegraphics[width=\columnwidth]{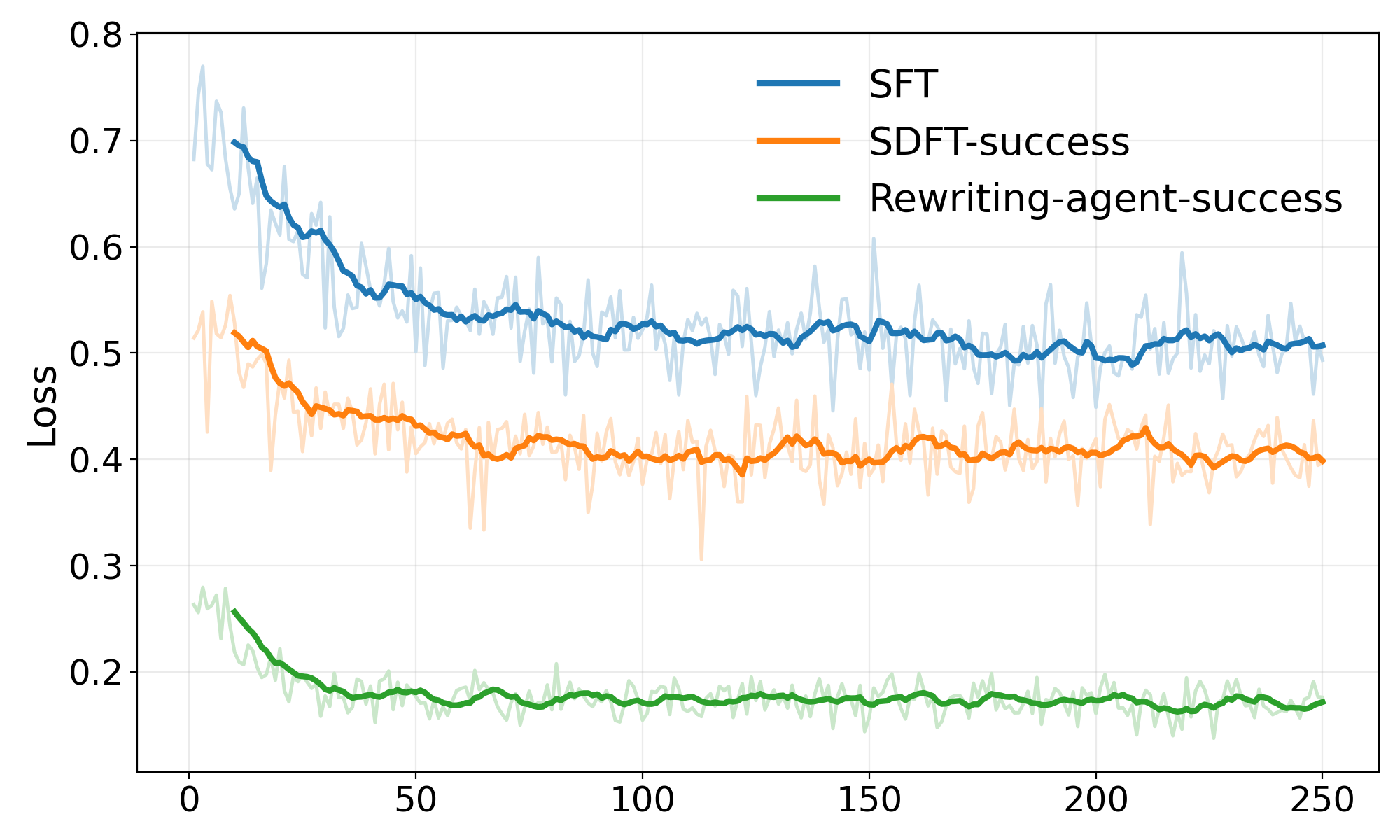}
  \caption{Downstream SFT training loss on the success-only subset for \texttt{Llama-3.2-3B-Instruct}.}
  \label{fig:loss_success}
\end{figure}

\subsection{Ablation Study}
We conduct an ablation study on \texttt{Llama-3.2-3B-Instruct} to quantify the contribution of each key component in our rewriting-agent framework.
Detailed per-benchmark ablation results for all variants are reported in Appendix~\ref{app:ablation-per-benchmark} (Table~\ref{tab:ablation-per-benchmark}).
Additional ablations on reward components, candidate group size $K$, the Generate--Verify--Fallback pipeline, hard gating, LoRA rank sensitivity, multi-seed robustness, and training cost breakdown are provided in Appendix~\ref{app:ablation}.

\paragraph{Summary of robustness analyses.}
The distribution-alignment reward has the largest effect on general-capability retention: removing it reduces GeneralAvg from 43.17 to 39.37, while removing the diversity reward reduces MathAvg from 21.12 to 20.18.
Hard gating outperforms soft shaping in both task-consistency yield (74.96 vs. 71.58) and MathAvg (21.12 vs. 19.52), and fallback restores full supervision coverage relative to success-only training.
Performance improves with LoRA rank and largely saturates at $r=64$; increasing $K$ shows diminishing returns; and the multi-seed results retain the same overall trend.

\paragraph{Success-only supervision quality.}
We conduct downstream SFT on the \emph{success-only} subset and report the training loss in Figure~\ref{fig:loss_success}, enabling a controlled comparison between rewrites that are successfully generated by our rewriting agent and those produced via successful self-rewriting (i.e., \texttt{SDFT}). Figure~\ref{fig:loss_success} shows that \texttt{Rewriting-agent-success} converges to a lower and smoother loss than \texttt{SDFT-success}, suggesting that our learned rewrites are more in-distribution and thus easier to fit under the QA-style maximum-likelihood objective. We provide qualitative case studies in Appendix~\ref{app:case_study}, showing that although the model’s direct rewrites (\texttt{SDFT}) are also task-correct, our learned rewrites better match the model’s natural QA-style completion patterns, thereby yielding a more stable and effective training signal.

\begin{table}[t]
  \centering
  \small
  \setlength{\tabcolsep}{3pt}
  \renewcommand{\arraystretch}{1.05}
  \begin{tabular}{lcccc}
    \toprule
    \textbf{Method} & \textbf{TaskAvg} & \textbf{Impr.} & \textbf{GenAvg} & \textbf{Impr.} \\
    \midrule
    \multicolumn{5}{l}{\textit{Logical Reasoning}} \\
    \midrule
    Instruct          & 45.03 & --      & 45.21 & --      \\
    SFT            & 73.80 & +63.89\% & 36.58 & $-$19.08\% \\
    SDFT           & 68.77 & +52.72\% & 39.44 & $-$12.76\% \\
    Mind the Gap   & 68.06 & +51.14\% & 39.67 & $-$12.25\% \\
    rewriting (MATH)  & 73.59 & +63.42\% & \textbf{43.32} & \textbf{$-$4.18\%} \\
    rewriting (logic) & \textbf{74.76} & \textbf{+66.02\%} & 43.31 & $-$4.20\% \\
    \midrule
    \multicolumn{5}{l}{\textit{Medical QA (MedQA)}} \\
    \midrule
    Instruct              & 50.24 & --         & 45.21 & --      \\
    SFT                   & \textbf{64.78} & \textbf{+28.94\%}   & 35.60 & $-$21.26\% \\
    SDFT                  & 58.45 & +16.34\%   & 38.44 & $-$14.97\% \\
    Mind the Gap          & 61.94 & +23.29\%   & 39.24 & $-$13.21\% \\
    rewriting (MATH)      & 63.85 & +27.09\%   & 43.22 & $-$4.40\% \\
    rewriting (med)       & 64.02 & +27.43\%   & \textbf{44.72} & \textbf{$-$1.08\%} \\
    \bottomrule
  \end{tabular}
  \caption{Preliminary cross-domain transfer on \texttt{Llama-3.2-3B-Instruct}: a math-trained rewriting agent applied to two unseen target domains.}
  \label{tab:cross-domain}
\end{table}

\begin{table}[t]
  \centering
  \small
  \setlength{\tabcolsep}{4pt}
  \renewcommand{\arraystretch}{1.05}
  \begin{tabular}{lcc}
    \toprule
    \textbf{Method} & \textbf{Logic (\%)} & \textbf{Medical (\%)} \\
    \midrule
    SDFT              & 56.71 & 62.12 \\
    Mind the Gap      & 58.93 & 63.64 \\
    rewriting (MATH)  & 68.31 & 70.48 \\
    rewriting (in-domain) & \textbf{70.21} & \textbf{71.01} \\
    \bottomrule
  \end{tabular}
  \caption{Task-consistency yield under cross-domain transfer.}
  \label{tab:cross-yield}
\end{table}

\paragraph{Cross-domain generalization.}
To test whether the rewriting agent learns a transferable rewriting strategy rather than domain-specific templates, we apply an agent trained exclusively on mathematical data to rewrite supervision in two distinct target domains.
(1)~\emph{Logical reasoning}: we use the publicly designated training split of \texttt{ProofWriter}~\citep{tafjord-etal-2021-proofwriter} as the training domain and evaluate on the held-out test split of \texttt{ProofWriter}, as well as \texttt{LogiQA-2.0}~\citep{liu2023logiqa} and \texttt{RuleTaker}~\citep{clark2020transformers}.
(2)~\emph{Medical QA}: we use the publicly designated training split of \texttt{MedQA}~\citep{jin2021disease} (medical licensing exam questions) as the downstream SFT corpus and evaluate only on its held-out test split. No test examples are used for rewriting-agent training or downstream SFT.
For each domain, we compare \texttt{rewriting (MATH)} (cross-domain transfer) with an in-domain upper bound that retrains the agent on the target domain.

As shown in Table~\ref{tab:cross-domain}, the cross-domain variant \texttt{rewriting (MATH)} achieves downstream task gains very close to the in-domain upper bound in both domains (73.59 vs.\ 74.76 on logic; 63.85 vs.\ 64.02 on medical), while its general-capability drop remains much smaller than vanilla SFT ($-$4.18\% vs.\ $-$19.08\% on logic; $-$4.40\% vs.\ $-$21.26\% on medical).
Table~\ref{tab:cross-yield} further reports the task-consistency yield: \texttt{rewriting (MATH)} achieves a substantially higher yield than existing rewriting baselines in both domains and remains close to the in-domain variant.
These results provide preliminary evidence that the rewriting policy transfers beyond its training domain for the same backbone, although broader validation is required before drawing general conclusions.
Full per-benchmark breakdowns are provided in Appendix~\ref{app:cross-domain-results}.

\section{Conclusion}

Downstream SFT effectively adapts instruction-tuned LLMs to target tasks, yet distribution mismatch between expert demonstrations and the model's generation distribution can intensify optimization instability and catastrophic forgetting. We propose an RL-based supervision-rewriting policy, implemented as lightweight LoRA patches on a frozen backbone and trained with GRPO using hard-gated, automatically evaluable rewards, to produce targets that are task-consistent, closer to the backbone's QA-style distribution, and diverse. Across the three evaluated backbones, our method achieves downstream performance broadly comparable to standard SFT while reducing general-domain degradation. Results on logic and medical QA further provide preliminary evidence of cross-domain reuse for the same backbone; broader validation across larger models and domains remains future work.

\section*{Limitations}
Our study has several limitations.
First, we evaluate the proposed rewriting-agent on instruction-tuned backbones with up to 7B parameters. While we expect the approach to scale to larger models, verifying this remains subject to computational constraints.
Second, our reward signals are fully automatic; incorporating human preference judgments could further refine rewriting quality.
Third, the current pipeline generates a single rewrite per instance. Exploring multi-candidate selection or iterative refinement may yield additional improvements.
Fourth, the reasoning-validity component of our automatic verifier has not been independently audited against human annotations. Errors from this verifier can therefore affect both policy training and rewritten-dataset construction, and a systematic human evaluation remains important future work.

\section*{Ethical Considerations}
Our method transforms training supervision and does not alter model inference-time safeguards. Nevertheless, automatic rewriting and verification can preserve or introduce errors and biases from the backbone, judge, or source data. Rewritten data should therefore be inspected before use in high-stakes domains, and deployment should retain domain-appropriate evaluation, monitoring, and safety controls. Our experiments use public research datasets and evaluate aggregate model performance; we do not collect personal or sensitive user data.

\bibliography{custom}

\appendix
\section{Use of Large Language Models (LLMs)}
In the preparation of this work, we used LLMs as auxiliary tools and as an explicitly disclosed component of the automatic verification pipeline. LLMs assisted in drafting portions of the code and in refining the wording of certain sentences for clarity and readability. In addition, DeepSeek-V3.2 was used as the reasoning-validity judge described in Sec.~\ref{sec:reward}; its binary judgments affect the task-consistency gate, rewriting rewards, and fallback-based dataset construction. All technical content, including the design of algorithms, experimental methodology, analysis, and interpretations, was independently developed by the authors. The judge's role is therefore part of the experimental method rather than a claim that LLM assistance had no influence on the reported results.

\section{Discussion: Relationship to Token-Level Probability Control}
\label{app:safety}

A natural question is whether our rewriting-based framework could interact with or undermine the safety alignment of the backbone model, given that it explicitly optimizes for higher token-level generatability under the model's distribution.
We clarify several important distinctions below.

\paragraph{Training-time data transformation, not inference-time intervention.}
Our method operates entirely at the \emph{data preparation} stage: the rewriting agent produces modified supervision targets offline, and the downstream model is then trained via standard SFT on these targets.
Unlike inference-time techniques such as logit patching, activation steering, or controlled decoding, our approach does not modify the model's decoding procedure or token-level probability distribution at test time.
The trained model is deployed with its standard autoregressive sampling, and no additional modules or probability manipulations are applied during inference.

\paragraph{Scope of distributional alignment.}
The distributional-alignment reward $r_{\text{dist}}$ encourages rewrites to be more generatable under the backbone's QA-style conditional distribution $\pi_0(\cdot \mid x)$.
This objective is designed to reduce the distribution mismatch associated with optimization instability and forgetting during SFT---it does not target or modify the model's safety-related behaviors (e.g., refusal boundaries, harmlessness constraints).
In principle, any post-training method that alters model behavior---including standard SFT, RLHF, and parameter-efficient fine-tuning---may inadvertently shift safety-aligned behaviors~\citep{qi2024finetuning}.
Our method is not exempt from this general concern, but it does not introduce a qualitatively new attack surface beyond what already exists in supervised fine-tuning pipelines.

\paragraph{Recommended safeguards.}
We emphasize that the intended use of our framework is to mitigate catastrophic forgetting during domain adaptation, not to circumvent safety mechanisms.
For deployment in safety-sensitive settings, we recommend applying standard post-training safety practices: (i)~systematic red-teaming of the rewritten data and the resulting fine-tuned model, (ii)~monitoring for refusal-boundary drift via safety benchmarks before and after fine-tuning, and (iii)~restricting the rewriting agent's training data to benign, task-relevant corpora.
These measures are consistent with best practices for any supervised fine-tuning pipeline and are not specific to our method.

\section{Why Standard SFT Can Be Unstable}
\label{app:sft_instability}
\paragraph{An importance-weighted view of distribution mismatch.}
Standard SFT minimizes the sequence-level negative log-likelihood on an expert dataset
$\mathcal D_E=\{(x,y^\star)\}$:
\begin{equation}
\mathcal L_{\text{SFT}}(\theta)
=
\mathbb{E}_{(x,y^\star)\sim \mathcal D_E}\!\left[-\log \pi_\theta(y^\star\mid x)\right],
\end{equation}
whose gradient is
\begin{equation}
\nabla_\theta \mathcal L_{\text{SFT}}(\theta)
=
\mathbb{E}_{(x,y^\star)\sim \mathcal D_E}\!\left[-\nabla_\theta \log \pi_\theta(y^\star\mid x)\right].
\end{equation}
This expectation is taken over a fixed expert dataset rather than samples from the current model
$\pi_\theta(\cdot\mid x)$. We use an importance-weighted form only as an
interpretive tool for understanding distribution mismatch; SFT itself is not
an off-policy reinforcement-learning algorithm.

Following prior work~\citep{wu2025generalizationsftreinforcementlearning}, the expert expectation can be
represented under the current policy by resampling and reweighting. Let $\mathcal D_x$ denote the empirical
marginal over inputs and let $p_E(y\mid x)$ denote the empirical conditional distribution of demonstrations
in $\mathcal D_E$. Assuming the support of $p_E(\cdot\mid x)$ is contained in that of
$\pi_\theta(\cdot\mid x)$, for $y\sim \pi_\theta(\cdot\mid x)$ we have
\begin{equation}
\begin{aligned}
&\mathbb{E}_{x\sim\mathcal D_x,\,y^\star\sim p_E(\cdot\mid x)}
 \left[-\nabla_\theta\log\pi_\theta(y^\star\mid x)\right] \\
&\quad=\mathbb{E}_{x\sim\mathcal D_x,\,y\sim\pi_\theta(\cdot\mid x)}
\left[\frac{p_E(y\mid x)}{\pi_\theta(y\mid x)}
\bigl(-\nabla_\theta\log\pi_\theta(y\mid x)\bigr)\right].
\end{aligned}
\end{equation}
For the common case of one empirical target per input, $p_E(y\mid x)$ is a point mass and the
numerator reduces to an indicator for the expert trajectory. This identity is useful for interpreting
the mismatch, but it does not turn SFT into an off-policy reinforcement-learning algorithm.

\paragraph{Importance-weight variance, not an extra SFT gradient factor.}
The resampling representation contains the inverse-probability ratio
$p_E(y\mid x)/\pi_\theta(y\mid x)$. If an expert target has very low probability under
$\pi_\theta$, an estimator based on current-policy samples observes that target rarely and assigns it a
large weight, producing a heavy-tailed, high-variance learning signal~\citep{metelli2020importance,jiang2020note}.
This observation should not be read as an additional $1/\pi_\theta$ factor in the ordinary expert-sample
SFT gradient, which is computed directly from the demonstrations. Rather, it provides an interpretable
signal for why substantial distribution mismatch can make optimization sensitive to the sampled targets.

\paragraph{Connection to policy drift and catastrophic forgetting (token-level intuition).}
Expanding the log-likelihood in an autoregressive, token-level form,
\begin{equation}
\log \pi_\theta(y^\star\mid x)
=
\sum_{t=1}^{|y^\star|}
\log \pi_\theta\!\left(y_t^\star \mid x, y^\star_{<t}\right),
\end{equation}
shows that an SFT step aggregates token-level updates over every demonstration position.
Under substantial distribution shift, these updates can move probability mass toward the downstream
examples and away from previously learned behaviors; repeated updates can therefore induce policy drift
and catastrophic forgetting~\citep{zhang2025onpolicyrlmeetsoffpolicy,zhu2025anchoredsupervisedfinetuning}.
The importance-weighted representation above gives a complementary high-variance intuition, but does not
claim that low-probability tokens create an unbounded ordinary SFT gradient.

\paragraph{Relevance to our approach.}
Motivated by this distribution-mismatch perspective, our method intervenes at the data source:
we learn a rewriting policy that adjusts supervision targets to be (i) task-consistent and
(ii) more generatable under the backbone's QA-style distribution $\pi_0(\cdot\mid x)$.
This data-level intervention is intended to reduce mismatch and the resulting sensitivity of downstream
SFT updates, which empirically helps stabilize optimization and mitigate forgetting.
\section{Dataset Details}
\label{app:dataset}

\subsection{Training Corpora}

\paragraph{NuminaMath-CoT.}
\texttt{NuminaMath-CoT} is a large-scale math reasoning corpus consisting of diverse competition- and school-level problems paired with chain-of-thought (CoT) solutions.
Its sources span Chinese K-12 exercises, AMC/AIME-style contests, and international Olympiad problems, collected primarily from online exam PDFs and mathematics discussion forums.
The dataset undergoes OCR, segmentation into problem--solution pairs, translation into English, and post-processing to align solutions into a CoT format with standardized final answers.
Each instance provides the problem statement and an English CoT solution (with a final answer), together with a coarse-grained source tag.
In our experiments, we sample from this corpus and apply a length filter to exclude overlong instances.

\paragraph{OpenMathReasoning.}
\texttt{OpenMathReasoning} is a large-scale math reasoning dataset built primarily from AoPS problems.
It contains multiple supervision modalities, including long CoT solutions and tool-integrated reasoning (TIR) solutions, as well as a selection set that chooses the most promising solution among candidates.
Problem statements are preprocessed for quality, and solutions are synthesized by strong open models.
We treat \texttt{OpenMathReasoning} as complementary training data to \texttt{NuminaMath-CoT} due to its high coverage of contest-style problems and its diverse solution traces.
In our setup, we merge the two corpora, randomly sample instances, and filter out examples whose tokenized length exceeds the model context limit.

\subsection{Math Evaluation Benchmarks}

\paragraph{Math500.}
\texttt{Math500} is a curated 500-problem subset of the \texttt{MATH} benchmark.
It features competition-style questions across multiple subjects (e.g., algebra, number theory, geometry, precalculus) and difficulty levels.
Each example provides a problem statement and a reference solution with a final answer string.
We evaluate by extracting the model's final answer and matching it against the reference answer under our normalization rules.

\paragraph{MinervaMath.}
\texttt{MinervaMath} targets advanced quantitative reasoning problems at the undergraduate level, covering broad STEM topics (e.g., physics/astronomy, basic engineering-style calculations, and mathematical modeling).
Problems typically require multi-step derivations and yield a short verifiable final answer (number or expression).
We use the official test split and report answer accuracy based on final-answer extraction.

\paragraph{AMC23.}
\texttt{AMC23} consists of problems from the 2023 AMC (American Mathematics Competitions), which are short, competition-style questions with concise numeric answers.
Compared with larger math benchmarks, AMC problems emphasize algebraic manipulation, counting/probability, and geometry in a compact format.
We report accuracy using final-answer matching.

\paragraph{AGIEval-Math.}
\texttt{AGIEval} is a suite of standardized-exam questions spanning multiple subjects and languages.
We define \texttt{AGIEval-Math} as the math-related subsets in \texttt{AGIEval}, including SAT-style math, AQuA-RAT quantitative reasoning, and Gaokao math in both QA and cloze-style formats, as well as the \texttt{MATH} subset included by AGIEval.
This benchmark mixes multiple-choice and open-form (cloze) questions; we follow the standard evaluation protocol for each subset and compute overall accuracy.

\paragraph{IMO-Bench (Answer subset).}
\texttt{IMO-Bench} targets Olympiad-level reasoning.
Since proof grading is expensive and often requires expert assessment, we focus on its answer-verifiable component (often referred to as \emph{IMO-AnswerBench}), which contains Olympiad problems with short, checkable final answers.
The problems are curated to cover diverse topics (algebra, combinatorics, geometry, number theory) and to reduce memorization via expert editing/robustification.
We report final-answer accuracy under the same extraction and normalization pipeline as other open-answer math benchmarks.

\subsection{General-Domain Benchmarks for Forgetting}

\paragraph{MMLU (math removed).}
\texttt{MMLU} is a general-domain multiple-choice benchmark covering a broad set of academic subjects.
To quantify catastrophic forgetting outside the math domain, we remove the math-related subjects (e.g., abstract algebra, college mathematics, elementary mathematics, high school mathematics, and high school statistics) and evaluate on the remaining subjects using the standard multiple-choice protocol.

\paragraph{MMLU-Pro (math removed).}
\texttt{MMLU-Pro} is a more challenging and carefully curated extension of \texttt{MMLU}, designed to reduce shortcut artifacts and increase difficulty.
We exclude the mathematics domain (and any explicitly math-labeled subsets) and report accuracy on the remaining domains using the standard multiple-choice protocol.

\paragraph{AGIEval (math removed).}
To measure general-domain transfer beyond math within the \texttt{AGIEval} suite, we exclude the math-related subsets used to form \texttt{AGIEval-Math} and evaluate on the remaining exam tasks (primarily multiple-choice) following the official evaluation procedure.

\section{Baseline Details}
\label{app:baselines}

\paragraph{Common setup.}
All baselines use the same backbone model $\pi_0$, the same task-consistency verifier, the same training/evaluation split, and identical downstream SFT hyperparameters. Let each training instance be $(x, y^\star)$, where $x$ is the instruction/input and $y^\star$ is the expert demonstration. For methods involving rewriting, we construct a training target $y$ via a unified \emph{verify--fallback} rule:
\[
y \;=\;
\begin{cases}
\tilde y, & \text{if } \mathrm{Verify}(x, y^\star, \tilde y)=1,\\
y^\star, & \text{otherwise},
\end{cases}
\]
where $\tilde y$ denotes the generated rewrite/candidate output.

\paragraph{\texttt{Vanilla SFT}.}
We directly perform supervised fine-tuning on the original corpus, i.e., $y = y^\star$ for every instance. No rewriting is applied.

\paragraph{\texttt{SDFT}~\citep{yang2024selfdistillationbridgesdistributiongap}.}
For each instance, we prompt the frozen backbone $\pi_0$ with a fixed rewriting template to produce a candidate rewrite $\tilde y$ from $(x, y^\star)$. We then apply the same task-consistency verifier. Verified rewrites are used as SFT targets; otherwise we fall back to the original expert demonstration $y^\star$.

\paragraph{\texttt{Mind the Gap}~\citep{zhao2025mindgapdatarewriting}.}
We first prompt $\pi_0$ to \emph{self-solve} the input $x$, producing a solution $\hat y$. If $\hat y$ passes verification, we keep it as the SFT target. Otherwise, we rewrite the expert demonstration using a fixed rewriting template to obtain $\tilde y$, verify it, and fall back to $y^\star$ if verification still fails. This procedure rewrites supervision only for the instances where self-solving fails.

\paragraph{\texttt{Rewriting-agent (Ours)}.}
We replace template-based rewriting with a learnable rewriting policy $R_\phi$ that generates a candidate rewrite $\tilde y = R_\phi(x, y^\star)$ (with $\pi_0$ as the backbone). We then construct the rewriting dataset using the same verify--fallback rule above and perform downstream SFT on the resulting targets. Compared to template-based baselines, this keeps the verification and fallback mechanism fixed while changing only the rewriting policy.

\section{Ablation Studies}
\label{app:ablation}

We conduct ablation studies on \texttt{Llama-3.2-3B-Instruct} to quantify the contribution of each key component.

\subsection{Reward Component Ablation}
\label{app:reward-ablation}

We ablate reward components in the Stage~I GRPO objective while keeping the backbone $\pi_0$, task-consistency verifier, data split, and downstream SFT hyperparameters fixed (Table~\ref{tab:ablation_reward}). Removing the distributional-alignment reward $r_{\text{dist}}$ causes the largest drop in generalization retention, indicating that encouraging QA-style generatability under $\pi_0(\cdot\mid x)$ is crucial for mitigating forgetting. Removing the diversity regularizer $r_{\text{div}}$ also degrades performance, with a more pronounced impact on in-domain math results. Notably, \texttt{task-only} attains the highest pass rate yet performs worst overall, showing that feasibility alone is insufficient without additional shaping toward in-distribution targets.

\begin{table}[h]
  \centering
  \small
  \setlength{\tabcolsep}{6pt}
  \renewcommand{\arraystretch}{1.1}
  \begin{tabular}{lccc}
    \toprule
    \textbf{Variant} & \textbf{MathAvg} & \textbf{GeneralAvg} & \textbf{TC-Yield} \\
    \midrule
    Full (ours)            & 21.12 & 43.17 & 74.96 \\
    w/o $r_{\text{dist}}$  & 20.73 & 39.37 & 74.58 \\
    w/o $r_{\text{div}}$   & 20.18 & 40.93 & 75.34 \\
    task-only              & 20.07 & 39.23 & 78.25 \\
    \bottomrule
  \end{tabular}
  \caption{Reward-component ablations of the Stage~I GRPO objective (TC-Yield in \%).}
  \label{tab:ablation_reward}
\end{table}

\subsection{Candidate Group Size $K$}
\label{app:k-sweep}

\begin{figure}[h]
  \centering
  \includegraphics[width=\columnwidth]{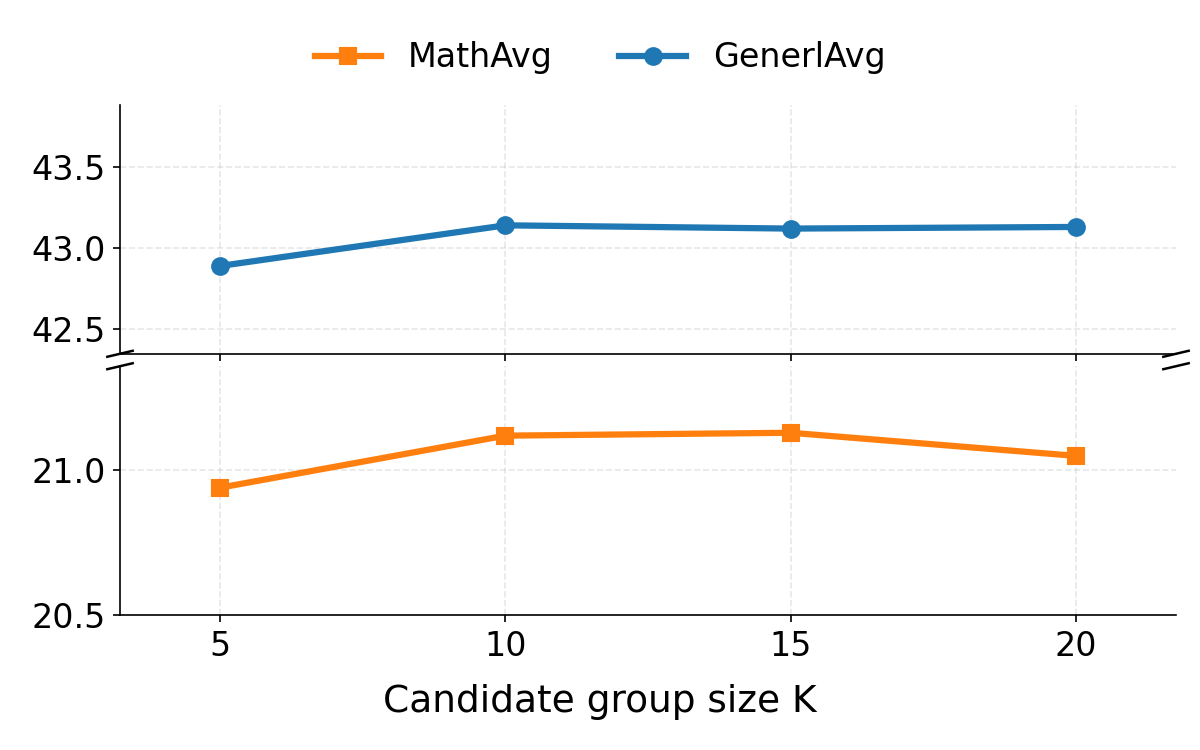}
  \caption{Effect of candidate group size $K$ in GRPO training of the rewriting agent on \texttt{Llama-3.2-3B-Instruct}.}
  \label{fig:k_sweep}
\end{figure}

We sweep $K\in\{5,10,15,20\}$ in GRPO training (Figure~\ref{fig:k_sweep}).
Larger $K$ generally improves performance but shows diminishing returns; we use $K{=}10$ by default as a practical trade-off.

\subsection{Generate--Verify--Fallback}
\label{app:additional-ablation}

We ablate the fallback step by training on \emph{success-only} rewrites and compare it with the default \textsc{Generate--Verify--Fallback} construction. As shown in Table~\ref{tab:fallback}, \emph{success-only} preserves general performance but limits downstream math gains due to reduced supervision coverage.

\begin{table}[h]
  \centering
  \small
  \setlength{\tabcolsep}{7pt}
  \renewcommand{\arraystretch}{1.1}
  \begin{tabular}{lccc}
    \toprule
    \textbf{Construction} & \textbf{Nums} & \textbf{MathAvg} & \textbf{GeneralAvg} \\
    \midrule
    Fallback (default) & 50000 & 21.12 & 43.17 \\
    Success-only       & 37483 & 20.30 & 43.70 \\
    \bottomrule
  \end{tabular}
  \caption{Ablation of fallback in dataset construction on \texttt{Llama-3.2-3B-Instruct}.
  Num. denotes the number of training instances used for downstream SFT.}
  \label{tab:fallback}
\end{table}

\subsection{Hard Gate vs.\ Soft Shaping}

We compare the proposed hard gating with a soft variant that always applies auxiliary rewards (Table~\ref{tab:ablation_gate}).
Soft shaping reduces PassGate (74.96$\rightarrow$71.58) and substantially hurts MathAvg (21.12$\rightarrow$19.52), suggesting that applying auxiliary rewards to incorrect candidates can introduce noisy shaping signals.
Hard gating yields a better overall trade-off by restricting auxiliary rewards to feasible rewrites.

\begin{table}[h]
  \centering
  \small
  \setlength{\tabcolsep}{6pt}
  \renewcommand{\arraystretch}{1.1}
  \begin{tabular}{lccc}
    \toprule
    \textbf{Variant} & \textbf{MathAvg} & \textbf{GeneralAvg} & \textbf{TC-Yield} \\
    \midrule
    Hard gate (ours) & 21.12 & 43.17 & 74.96 \\
    Soft shaping     & 19.52 & 42.96 & 71.58 \\
    \bottomrule
  \end{tabular}
  \caption{Hard gating vs.\ soft shaping on \texttt{Llama-3.2-3B-Instruct}.}
  \label{tab:ablation_gate}
\end{table}

\subsection{LoRA Rank Sensitivity}
\label{app:lora-rank}

We vary the LoRA configuration $(r, \alpha)$ (with $\alpha = r$) while keeping other settings fixed on \texttt{Llama-3.2-3B-Instruct} (Table~\ref{tab:lora-rank}).
Performance improves monotonically with larger rank and largely saturates around $r \approx 64$, suggesting the method is not overly sensitive to a single patch-strength choice.

\begin{table}[h]
  \centering
  \small
  \setlength{\tabcolsep}{6pt}
  \renewcommand{\arraystretch}{1.1}
  \begin{tabular}{lcc}
    \toprule
    \textbf{Setting} & \textbf{MathAvg} & \textbf{GeneralAvg} \\
    \midrule
    $r{=}16,\;\alpha{=}16$ & 19.81 & 40.26 \\
    $r{=}32,\;\alpha{=}32$ & 20.87 & 42.32 \\
    $r{=}64,\;\alpha{=}64$ & 21.12 & 43.17 \\
    $r{=}128,\;\alpha{=}128$ & 21.14 & 43.19 \\
    \bottomrule
  \end{tabular}
  \caption{Effect of LoRA rank on rewriting-agent performance (\texttt{Llama-3.2-3B-Instruct}).}
  \label{tab:lora-rank}
\end{table}

\subsection{Multi-Seed Results}
\label{app:multi-seed}

To verify robustness, we rerun key experiments with multiple random seeds and report mean $\pm$ standard error (Table~\ref{tab:multi-seed}).
The improvements of our method remain consistent across seeds, indicating that the gains are not driven by a single favorable random seed.

\begin{table*}[h]
  \centering
  \small
  \setlength{\tabcolsep}{4pt}
  \renewcommand{\arraystretch}{1.15}
  \begin{tabular}{l cccccccc}
    \toprule
    \textbf{Model}
      & \textbf{AMC23} & \textbf{AGI-Math} & \textbf{IMO} & \textbf{Math500} & \textbf{MinervaMath}
      & \textbf{MMLU} & \textbf{MMLU-Pro} & \textbf{AGI} \\
    \midrule
    Llama-3.2-3B & 17.63{\tiny$\pm$0.32} & 30.28{\tiny$\pm$0.44} & 5.84{\tiny$\pm$0.37} & 33.05{\tiny$\pm$0.41} & 7.61{\tiny$\pm$0.35} & 61.18{\tiny$\pm$0.48} & 35.04{\tiny$\pm$0.39} & 39.32{\tiny$\pm$0.46} \\
    + SFT & 24.91{\tiny$\pm$0.45} & 32.81{\tiny$\pm$0.33} & 5.65{\tiny$\pm$0.41} & 36.88{\tiny$\pm$0.36} & 8.71{\tiny$\pm$0.40} & 48.15{\tiny$\pm$0.31} & 31.12{\tiny$\pm$0.47} & 32.56{\tiny$\pm$0.38} \\
    + SDFT & 20.14{\tiny$\pm$0.34} & 32.18{\tiny$\pm$0.49} & 5.97{\tiny$\pm$0.32} & 33.12{\tiny$\pm$0.44} & 8.16{\tiny$\pm$0.36} & 55.11{\tiny$\pm$0.43} & 31.08{\tiny$\pm$0.35} & 34.69{\tiny$\pm$0.41} \\
    + Mind the Gap & 22.44{\tiny$\pm$0.38} & 31.18{\tiny$\pm$0.40} & 6.52{\tiny$\pm$0.46} & 33.29{\tiny$\pm$0.33} & 8.13{\tiny$\pm$0.49} & 54.21{\tiny$\pm$0.37} & 31.47{\tiny$\pm$0.31} & 35.61{\tiny$\pm$0.44} \\
    + GRPO & 20.87{\tiny$\pm$0.94} & 30.41{\tiny$\pm$0.43} & 6.18{\tiny$\pm$0.43} & 34.46{\tiny$\pm$0.32} & 7.63{\tiny$\pm$0.48} & 57.96{\tiny$\pm$0.25} & 33.86{\tiny$\pm$0.42} & 39.48{\tiny$\pm$0.83} \\
    + Rewriting-agent & 23.87{\tiny$\pm$0.41} & 33.41{\tiny$\pm$0.36} & 6.88{\tiny$\pm$0.34} & 35.46{\tiny$\pm$0.48} & 8.63{\tiny$\pm$0.42} & 58.96{\tiny$\pm$0.39} & 33.91{\tiny$\pm$0.45} & 39.18{\tiny$\pm$0.33} \\
    \bottomrule
  \end{tabular}
  \caption{Multi-seed results (mean $\pm$ standard error) on \texttt{Llama-3.2-3B-Instruct}.}
  \label{tab:multi-seed}
\end{table*}

\subsection{Training Cost Breakdown}
\label{app:cost}

Table~\ref{tab:cost} reports the compute cost (GPU-hours on A100-80G) for each stage of the pipeline on \texttt{Llama-3.2-3B-Instruct}.
The extra cost of our method is concentrated in agent training, which is amortizable: once trained, the rewriting agent can be reused across domains without retraining.

\begin{table*}[h]
  \centering
  \small
  \setlength{\tabcolsep}{8pt}
  \renewcommand{\arraystretch}{1.1}
  \begin{tabular}{lccccc}
    \toprule
    \textbf{Method} & \textbf{Agent} & \textbf{Rewrite} & \textbf{Train} & \textbf{MathAvg} & \textbf{GenAvg} \\
    \midrule
    SFT            & --     & --    & 5.529  & 21.81 & 37.26 \\
    SDFT           & --     & 1.357 & 5.743  & 19.84 & 40.32 \\
    Mind the Gap   & --     & 2.403 & 5.672  & 20.30 & 40.44 \\
    Ours           & 38.226 & 1.472 & 5.811  & 21.12 & 43.17 \\
    GRPO           & --     & --    & 33.368 & 19.83 & 43.18 \\
    \bottomrule
  \end{tabular}
  \caption{Compute breakdown (GPU-hours, A100-80G) and performance on \texttt{Llama-3.2-3B-Instruct}. Agent cost is one-time and amortizable via cross-domain reuse.}
  \label{tab:cost}
\end{table*}

\section{Training Details}
\label{app:training-details}

\subsection{Downstream SFT}
We perform downstream supervised fine-tuning (SFT) with \texttt{LLaMA-Factory}~\citep{zheng2024llamafactory} using the AdamW optimizer and bfloat16 mixed precision.
Unless otherwise specified, we use the default AdamW hyperparameters (i.e., $\beta_1{=}0.9$, $\beta_2{=}0.999$, $\epsilon{=}10^{-8}$), apply gradient clipping with max norm 1.0, and set weight decay to 0.0.
For all backbones, we fine-tune for 2 epochs with a per-device batch size of 4 and gradient accumulation of 8, resulting in a global batch size of 128 when using 4 GPUs.
We adopt a cosine learning-rate schedule with a warmup ratio of 0.1.
We use a learning rate of $5\times10^{-6}$ for \texttt{Mistral-7B-Instruct-v0.3} and $7\times10^{-6}$ for both \texttt{Llama-3.2-1B-Instruct} and \texttt{Llama-3.2-3B-Instruct}.
To support full-parameter fine-tuning under limited GPU memory, we enable DeepSpeed ZeRO-3 with the configuration file \texttt{ds\_z3\_config.json} (provided in the \texttt{LLaMA-Factory} examples), which shards optimizer states, gradients, and model parameters across GPUs.
We run ZeRO-3 without CPU offloading.
Training is executed with distributed data parallelism; we set a sufficiently large DDP timeout to avoid premature termination in long-sequence runs.

\subsection{Rewriting-agent training (GRPO + LoRA)}
We train the rewriting agent with \texttt{verl}, an RL training framework for LLMs.
The instruction-tuned backbone $\pi_0$ is frozen, and we parameterize the rewriting policy $R_{\phi}$ using LoRA patch adapters applied to all linear layers, with rank $r{=}64$ and $\alpha{=}64$.
We optimize only the LoRA parameters using GRPO-style policy optimization.
For GRPO training, we use a global batch of 512 prompts and sample $K$ candidate rewrites per prompt for group-relative updates (default $K{=}10$, unless explicitly swept).
We cap the maximum prompt length and response length to 2048 and 4096 tokens, respectively, and filter or truncate overlong prompts following the \texttt{verl} data pipeline.

To accelerate rollouts and log-probability computation, we use a \texttt{vLLM}-based rollout engine during GRPO training, with conservative GPU memory utilization to stabilize long-sequence decoding.
Reward computation uses our custom automatically evaluable signals (task-consistency gate, QA-style generatability under $\pi_0(\cdot\mid x)$, and within-group diversity) as described in Sec.~\ref{sec:reward}.
For embedding-based similarity computations in the reward (e.g., the diversity term), we use the \texttt{Qwen-Embedding-0.6B} model.

\subsection{Rewriting inference with vLLM}
To generate rewritten datasets from the trained agent, we perform batched decoding with \texttt{vLLM}~\citep{kwon2023efficientmemorymanagementlarge} using tensor parallelism over 4 GPUs.
We use greedy decoding for determinism (temperature $=0.0$, $top\_p=1.0$) and allow up to 8192 new tokens per output, without specifying an explicit stop sequence.
We use a batch size of 36 for generation and do not apply retries for failed rewrites (failed cases are handled by verification and fallback during dataset construction).

\subsection{Prompt Templates}
\label{app:prompt-templates}
We use the following prompt templates for rewriting, math evaluation, and general multiple-choice evaluation.

\begin{promptbox}{Rewriting Prompt}
\begin{lstlisting}[style=promptstyle]
You are an expert in math word problems. Below is a math word problem and an existing step-by-step solution.
Please rewrite the solution in your own words while keeping all reasoning steps correct and keeping the final answer the same.
Follow these rules:
- Keep the explanation clear and step by step.
- Do NOT mention that you are rewriting another solution.
- At the end, on a separate line, output ONLY the final answer in LaTeX boxed format, exactly like:
$\boxed{56}$
Use dollar signs and \boxed{} exactly as shown.

Problem:
{question}

Existing solution (for reference, do NOT copy it verbatim):
{original_solution}
\end{lstlisting}
\end{promptbox}

\begin{promptbox}{Math Evaluation Prompt}
\begin{lstlisting}[style=promptstyle]
You are an expert in math word problems.
Read the following problem and solve it step by step.
At the end, on a separate line, output ONLY the final answer in LaTeX boxed format, exactly like:
$\boxed{56}$
Use dollar signs and \boxed{} exactly as shown.
Problem: {question}
\end{lstlisting}
\end{promptbox}

\begin{promptbox}{General Evaluation Prompt}
\begin{lstlisting}[style=promptstyle]
You are an expert problem solver.
Read the following multiple-choice question and solve it step by step.
Choose the best option.
At the end, on a separate line, output ONLY the final option letter in LaTeX boxed format, exactly like:
$\boxed{A}$
Use dollar signs and \boxed{} exactly as shown.
Question: {question}
Options:
{options}
\end{lstlisting}
\end{promptbox}

\section{Case Study}
\label{app:case_study}
\lstset{literate={÷}{{$\div$}}1}
\begin{promptbox}{Example Problem}
\begin{lstlisting}[style=promptstyle]
What is the greatest 3-digit base 8 positive integer that is divisible by 7?
(Express your answer in base 8.)
\end{lstlisting}
\end{promptbox}

\begin{promptbox}{Original Demonstration (loss $\approx 0.74$)}
\begin{lstlisting}[style=promptstyle]
1. The greatest 3-digit number in base 8 is $777_8$.
2. Convert $777_8$ to decimal:
   \[
   7 \cdot 8^2 + 7 \cdot 8^1 + 7 \cdot 8^0 = 7 \cdot 64 + 7 \cdot 8 + 7 = 448 + 56 + 7 = 511
   \]
3. Find the largest number less than or equal to 511 that is divisible by 7. The closest number is:
   \[
   511 \div 7 = 73 \text{ R } 0
   \]
   Therefore, 511 itself is divisible by 7.
4. Convert 511 back to base 8:
   - $511 \div 8 = 63$ remainder 7
   - $63 \div 8 = 7$ remainder 7
   - $7 \div 8 = 0$ remainder 7
   Thus, $511_{10} = 777_8$.

Conclusion:
The greatest 3-digit base 8 number divisible by 7 is $\boxed{777_8}$.
\end{lstlisting}
\end{promptbox}

\begin{promptbox}{Direct Rewrite (loss $\approx 0.63$)}
\begin{lstlisting}[style=promptstyle]
To find the greatest 3-digit base 8 positive integer that is divisible by 7, we first identify the largest 3-digit number in base 8, which is $777_8$.

We then convert $777_8$ to decimal to check for divisibility by 7. The conversion involves calculating $7 \cdot 8^2 + 7 \cdot 8^1 + 7 \cdot 8^0$, which simplifies to $7 \cdot 64 + 7 \cdot 8 + 7 = 448 + 56 + 7 = 511$.

Since $511$ is divisible by 7 with no remainder, it is the number we are looking for. To express $511$ in base 8, we perform successive divisions by 8 and record the remainders: $511 \div 8 = 63$ remainder 7, $63 \div 8 = 7$ remainder 7, and $7 \div 8 = 0$ remainder 7. Therefore, $511_{10} = 777_8$.

The greatest 3-digit base 8 number divisible by 7 is $\boxed{777_8}$.
\end{lstlisting}
\end{promptbox}

\begin{promptbox}{Ours (loss $\approx 0.44$)}
\begin{lstlisting}[style=promptstyle]
Step 1: Understand the problem
We are asked to find the greatest 3-digit base 8 positive integer that is divisible by 7.

Step 2: Find the greatest 3-digit base 8 number
The greatest 3-digit number in base 8 is $777_8$.

Step 3: Convert $777_8$ to decimal
To convert $777_8$ to decimal, we need to multiply each digit by the corresponding power of 8 and add them together.
$7 \cdot 8^2 + 7 \cdot 8^1 + 7 \cdot 8^0 = 7 \cdot 64 + 7 \cdot 8 + 7 = 448 + 56 + 7 = 511$

Step 4: Find the largest number less than or equal to 511 that is divisible by 7
To find the largest number less than or equal to 511 that is divisible by 7, we can divide 511 by 7.
$511 \div 7 = 73$ with a remainder of 0, so 511 is divisible by 7.

Step 5: Convert 511 back to base 8
To convert 511 back to base 8, we can divide it by 8 and keep track of the remainders.
$511 \div 8 = 63$ with a remainder of 7
$63 \div 8 = 7$ with a remainder of 7
$7 \div 8 = 0$ with a remainder of 7
So, $511_{10} = 777_8$

Step 6: Determine the greatest 3-digit base 8 number divisible by 7
Since 511 is the greatest 3-digit base 8 number that is divisible by 7, the greatest 3-digit base 8 number divisible by 7 is $777_8$.

$\boxed{777_8}$
\end{lstlisting}
\end{promptbox}

\noindent\textbf{Brief Analysis.}
This example is task-correct across all versions, but the supervision targets differ in how learnable and distribution-aligned they are under SFT.
\begin{itemize}\setlength{\itemsep}{2pt}
  \item \textbf{Original (0.74)} is correct but includes more verbose narration and repeated conversion scaffolding, which increases sequence length and token-level uncertainty.
  \item \textbf{Direct rewrite (0.63)} reduces redundancy, improving conciseness and slightly lowering loss, but still follows a generic narrative template.
  \item \textbf{Ours (0.44)} yields the lowest loss: the rewritten target retains only the essential dependencies needed to justify the answer, while matching the model’s preferred structured problem-solving format, resulting in a sharper training signal.
\end{itemize}

\section{Details Results}

\subsection{Per-benchmark Breakdown of Main Results}
\label{app:combined-results}
\begin{table*}[!t]
  \centering
  \small
  \setlength{\tabcolsep}{5pt}
  \renewcommand{\arraystretch}{1.15}
  \begin{tabular}{l ccccc ccc}
    \toprule
    \multirow{2}{*}{\textbf{Model}}
      & \multicolumn{5}{c}{\textbf{Math Benchmarks}}
      & \multicolumn{3}{c}{\textbf{Generalization}} \\
    \cmidrule(lr){2-6}\cmidrule(lr){7-9}
      & \textbf{AMC23} & \textbf{AGI-Math}
      & \textbf{IMO} & \textbf{Math500} & \textbf{MinervaMath}
      & \textbf{MMLU} & \textbf{MMLU-Pro} & \textbf{AGI} \\
    \midrule

    \textbf{Llama-3.2-1B-Instruct}
      & 15.00 & 19.55 & 2.63 & 22.00 & 2.94
      & 38.08 & 17.55 & 26.39 \\
    \quad + \texttt{SFT}
      & 17.50 & 26.82 & 5.26 & 23.40 & 4.04
      & 33.92 & 12.51 & 21.37 \\
    \quad + \texttt{SDFT}
      & 17.50 & 21.36 & 3.51 & 23.20 & 3.44
      & 34.02 & 14.21 & 23.17 \\
    \quad + \texttt{Mind the Gap}
      & 15.00 & 20.18 & 3.51 & 22.60 & 3.31
      & 34.83 & 15.09 & 23.32 \\
    \quad + \texttt{GRPO}
      & 15.00 & 20.32 & 3.44 & 22.32 & 3.32
      & 37.44 & 14.63 & 25.47 \\
    \quad + \texttt{Rewriting-agent}
      & 17.50 & 25.45 & 4.39 & 24.20 & 3.94
      & 36.68 & 15.94 & 25.14 \\
    \addlinespace

    \textbf{Llama-3.2-3B-Instruct}
      & 17.50 & 30.36 & 5.70 & 33.20 & 7.72
      & 61.07 & 35.11 & 39.45 \\
    \quad + \texttt{SFT}
      & 25.00 & 32.72 & 5.70 & 36.80 & 8.82
      & 48.26 & 31.06 & 32.45 \\
    \quad + \texttt{SDFT}
      & 20.00 & 32.27 & 5.82 & 33.00 & 8.09
      & 55.19 & 31.02 & 34.75 \\
    \quad + \texttt{Mind the Gap}
      & 22.50 & 31.09 & 6.41 & 33.40 & 8.09
      & 54.12 & 31.53 & 35.68 \\
    \quad + \texttt{GRPO}
      & 20.00 & 30.72 & 6.14 & 34.60 & 7.72
      & 57.44 & 33.63 & 38.47 \\
    \quad + \texttt{Rewriting-agent}
      & 22.50 & 33.18 & 6.65 & 35.00 & 8.25
      & 58.51 & 33.44 & 37.55 \\
    \addlinespace
    
    \textbf{Mistral-7B-Instruct-v0.3}
      & 10.00 & 22.73 & 3.95 & 11.20 & 5.15
      & 52.14 & 29.52 & 39.95 \\
    \quad + \texttt{SFT}
      & 10.00 & 31.82 & 5.26 & 22.40 & 7.35
      & 38.83 & 23.52 & 35.94 \\
    \quad + \texttt{SDFT}
      & 12.50 & 24.55 & 5.14 & 20.40 & 6.25
      & 44.77 & 24.10 & 37.21 \\
    \quad + \texttt{Mind the Gap}
      & 15.00 & 25.00 & 5.26 & 20.20 & 7.56
      & 43.71 & 23.18 & 36.41 \\
    \quad + \texttt{GRPO}
      & 10.00 & 25.31 & 5.71 & 19.42 & 7.32
      & 47.12 & 23.25 & 38.47 \\
    \quad + \texttt{Rewriting-agent}
      & 12.50 & 32.45 & 7.45 & 21.40 & 8.56
      & 47.68 & 26.86 & 38.12 \\
    \bottomrule
  \end{tabular}
  \caption{Per-benchmark main results (accuracy, \%). Math benchmarks evaluate in-domain mathematical reasoning. Generalization benchmarks are reported on math-removed subsets (higher is better).}
  \label{tab:combined-results}
\end{table*}

\noindent This appendix reports the full per-benchmark accuracies (\%) corresponding to the aggregated scores in Table~\ref{tab:summary-results}.
We evaluate five in-domain math benchmarks (AMC23, AGI-Math, IMO, Math500, MinervaMath) and three out-of-domain generalization benchmarks (MMLU, MMLU-Pro, AGI), where math-related subsets are removed for the generalization suites as described in \S\ref{exp:datasets}.
All results are reported as accuracy (\%), and higher is better.

\subsection{Per-benchmark Cross-Domain Results}
\label{app:cross-domain-results}

Table~\ref{tab:cross-domain-logic-full} reports the full per-benchmark results for the cross-domain logical reasoning experiment (corresponding to Table~\ref{tab:cross-domain} in the main text).
Table~\ref{tab:cross-domain-med-full} reports the full per-benchmark results for the cross-domain medical experiment (corresponding to Table~\ref{tab:cross-domain} in the main text).

\begin{table*}[h]
  \centering
  \small
  \setlength{\tabcolsep}{5pt}
  \renewcommand{\arraystretch}{1.15}
  \begin{tabular}{l ccc ccc}
    \toprule
    \multirow{2}{*}{\textbf{Method}}
      & \multicolumn{3}{c}{\textbf{Logic Benchmarks}}
      & \multicolumn{3}{c}{\textbf{Generalization}} \\
    \cmidrule(lr){2-4}\cmidrule(lr){5-7}
      & \textbf{LogiQA-2.0} & \textbf{RuleTaker} & \textbf{ProofWriter}
      & \textbf{MMLU} & \textbf{MMLU-Pro} & \textbf{AGI} \\
    \midrule
    Llama-3.2-3B          & 47.70 & 34.50 & 52.90 & 61.07 & 35.11 & 39.45 \\
    \quad + SFT            & 51.40 & 72.60 & 97.40 & 52.30 & 26.80 & 30.63 \\
    \quad + SDFT           & 49.22 & 67.70 & 89.40 & 54.47 & 29.61 & 34.25 \\
    \quad + Mind the Gap   & 49.35 & 66.32 & 88.52 & 54.89 & 30.38 & 33.74 \\
    \quad + rewriting (MATH)  & 51.17 & 74.40 & 95.20 & 57.96 & 33.25 & 38.74 \\
    \quad + rewriting (logic) & 52.05 & 73.78 & 98.45 & 58.42 & 33.55 & 37.98 \\
    \bottomrule
  \end{tabular}
  \caption{Per-benchmark cross-domain results: logical reasoning on \texttt{Llama-3.2-3B-Instruct} (accuracy, \%).}
  \label{tab:cross-domain-logic-full}
\end{table*}

\begin{table*}[h]
  \centering
  \small
  \setlength{\tabcolsep}{5pt}
  \renewcommand{\arraystretch}{1.15}
  \begin{tabular}{l c ccc}
    \toprule
    \multirow{2}{*}{\textbf{Method}}
      & \multicolumn{1}{c}{\textbf{Medical}}
      & \multicolumn{3}{c}{\textbf{Generalization}} \\
    \cmidrule(lr){2-2}\cmidrule(lr){3-5}
      & \textbf{MedQA}
      & \textbf{MMLU} & \textbf{MMLU-Pro} & \textbf{AGI} \\
    \midrule
    Llama-3.2-3B            & 50.24 & 61.07 & 35.11 & 39.45 \\
    \quad + SFT              & 64.78 & 51.94  & 25.43 & 29.43 \\
    \quad + SDFT             & 58.45 & 53.24 & 29.81 & 32.28 \\
    \quad + Mind the Gap     & 61.94 & 54.32  & 30.11  & 33.28 \\
    \quad + rewriting (MATH) & 63.85 & 59.32  & 32.12 & 38.23 \\
    \quad + rewriting (med)  & 64.02 & 60.83 & 34.12  & 39.21\\
    \bottomrule
  \end{tabular}
  \caption{Per-benchmark cross-domain results: medical domain (\texttt{MedQA}) on \texttt{Llama-3.2-3B-Instruct} (accuracy, \%).}
  \label{tab:cross-domain-med-full}
\end{table*}

\subsection{Per-benchmark Ablation Results}
\label{app:ablation-per-benchmark}

Table~\ref{tab:ablation-per-benchmark} provides a per-benchmark breakdown of all ablation variants on \texttt{Llama-3.2-3B-Instruct}.
We report accuracy (\%) on downstream math benchmarks and on general-domain benchmarks with math-related subsets removed.
Unless otherwise specified, variants use the same verification protocol and dataset construction procedure as in the main setting; \texttt{Success-only} trains downstream SFT using only rewrites that pass the task-consistency gate.

\begin{table*}[!t]
  \centering
  \small
  \setlength{\tabcolsep}{7pt}
  \renewcommand{\arraystretch}{1.15}
  \begin{tabular}{l ccccc ccc}
    \toprule
    \multirow{2}{*}{\textbf{Variant}}
      & \multicolumn{5}{c}{\textbf{Math Benchmarks}}
      & \multicolumn{3}{c}{\textbf{Generalization (math-removed)}} \\
    \cmidrule(lr){2-6}\cmidrule(lr){7-9}
      & \textbf{AMC23} & \textbf{AGI-Math}
      & \textbf{IMO} & \textbf{Math500} & \textbf{MinervaMath}
      & \textbf{MMLU} & \textbf{MMLU-Pro} & \textbf{AGI} \\
    \midrule
     Rewriting-agent 
      & 22.50 & 33.18 & 6.65 & 35.00 & 8.25
      & 58.51 & 33.44 & 37.55 \\
     w/o $r_{\text{dist}}$
      & 22.50 & 32.86 & 6.15 & 34.20 & 7.93
      & 53.14 & 31.20 & 33.76 \\
     w/o $r_{\text{div}}$   
      & 20.00 & 32.17 & 6.43 & 34.00 & 8.31
      & 55.13 & 32.13 & 35.54 \\
    task-only    
      & 20.00 & 32.46 & 6.12 & 33.60 & 7.84
      & 53.10 & 30.42 & 34.19 \\
    Soft shaping 
      & 17.25 & 32.14 & 6.01 & 34.24 & 7.94
      & 58.34 & 32.81 & 37.72 \\
    Success-only
      & 22.50 & 31.09 & 6.41 & 33.40 & 8.08
      & 58.82 & 33.53 & 38.75 \\
    \bottomrule
  \end{tabular}
  \caption{Per-benchmark ablation results on \texttt{Llama-3.2-3B-Instruct} (accuracy, \%). Math benchmarks evaluate downstream mathematical reasoning; generalization benchmarks are evaluated on math-removed subsets. Higher is better.}
  \label{tab:ablation-per-benchmark}
\end{table*}

\end{document}